\RequirePackage{fix-cm}

\documentclass[twocolumn]{svjour3}

\smartqed  % flush right qed marks, e.g. at end of proof

\usepackage{graphicx}
\usepackage{subfig}
\usepackage{amsmath}
\usepackage{url}
\usepackage{booktabs}
\usepackage{threeparttable}
\usepackage{algorithm2e}
\graphicspath{{images/},{images/troubleshooting/}}
\usepackage[numbers, compress]{natbib}
\usepackage{tikz}

\newcommand{\ra}[1]{\renewcommand{\arraystretch}{#1}}
\newcommand{\specialcell}[2][c]{%
  \begin{tabular}[#1]{@{}l@{}}#2\end{tabular}}

\begin{document}

\title{Fixed-sized representation learning from Offline Handwritten Signatures of different sizes}
%\subtitle{Do you have a subtitle?\\ If so, write it here}

%\titlerunning{Short form of title}        % if too long for running head

\author{Luiz G Hafemann         \and
		Luiz S Oliveira			\and
        Robert Sabourin
}

%\authorrunning{Short form of author list} % if too long for running head

\institute{L. Hafemann \and R. Sabourin \at
              Laboratoire d'imagerie, de vision et d'intelligence artificielle (LIVIA) \\
			  \'Ecole de technologie sup\'erieure, 1100, rue Notre-Dame Ouest, \\
			  Montreal, QC, H3C 1K3, Canada \\
              \email{lghafemann@livia.etsmtl.ca}           %  \\
%             \emph{Present address:} of F. Author  %  if needed
           \and L. Oliveira \at
Department of Informatics \\
Federal University of Parana (UFPR), Rua Cel. Francisco Heraclito dos Santos, 100, Curitiba, PR, Brazil
}

\date{Received: date / Accepted: date}
% The correct dates will be entered by the editor

\usetikzlibrary{calc}
\maketitle
%\copyrightnotice
\begin{tikzpicture}[remember picture, overlay]
\node at ($(current page.south) + (0,+0.6in)$)[align=left] {This is a pre-print of an article published in the International Journal on Document Analysis and Recognition.\\
	
The final authenticated version is available online at: \url{https://doi.org/10.1007/s10032-018-0301-6}.};
\end{tikzpicture}

\begin{abstract}
Methods for learning feature representations for Offline Handwritten Signature Verification have been successfully proposed in recent literature, using Deep Convolutional Neural Networks to learn representations from signature pixels. Such methods reported large performance improvements compared to handcrafted feature extractors. However, they also introduced an important constraint: the inputs to the neural networks must have a fixed size, while signatures vary significantly in size between different users. In this paper we propose addressing this issue by learning a fixed-sized representation from variable-sized signatures by modifying the network architecture, using Spatial Pyramid Pooling. We also investigate the impact of the resolution of the images used for training, and the impact of adapting (fine-tuning) the representations to new operating conditions (different acquisition protocols, such as writing instruments and scan resolution). On the GPDS dataset, we achieve results comparable with the state-of-the-art, while removing the constraint of having a maximum size for the signatures to be processed. We also show that using higher resolutions (300 or 600dpi) can improve performance when skilled forgeries from a subset of users are available for feature learning, but lower resolutions (around 100dpi) can be used if only genuine signatures are used. Lastly, we show that fine-tuning can improve performance when the operating conditions change.

% \PACS{PACS code1 \and PACS code2 \and more}
% \subclass{MSC code1 \and MSC code2 \and more}
\end{abstract}

\section{Introduction}
\label{intro}

The handwritten signature is a behavioral biometric trait that is extensively used to verify a person's identity in legal, financial and administrative  areas. Automating the verification of handwritten signatures has been a subject of research since the decade of 1970 \cite{plamondon_automatic_1989, leclerc_automatic_1994, impedovo_automatic_2008, hafemann_offline_2015}, considering two scenarios: online (dynamic) and offline (static). In the online case, signatures are captured using a special device, such as a pen tablet, that records the dynamic information of the writing process (e.g. position of the pen over time). For offline signature verification, we consider signatures written on paper, that are subsequently scanned to be used as input. 

Most of the research effort in offline signature verification has been devoted to finding good feature representations for signatures, by proposing new feature descriptors for the problem \cite{hafemann_offline_2015}. Recent work, however, showed that learning features from data (signature images) can improve system performance to a large extent \cite{hafemann_ijcnn_2016, hafemann_pr_2017, rantzsch_signature_2016, zhang_multi-phase_2016}. These work rely on training Deep Convolutional Neural Networks (CNNs) to learn a hierarchy of representations directly from signature pixels. 

\begin{figure}
\centering
\subfloat[Directly resizing signatures]{\includegraphics[scale=0.28]{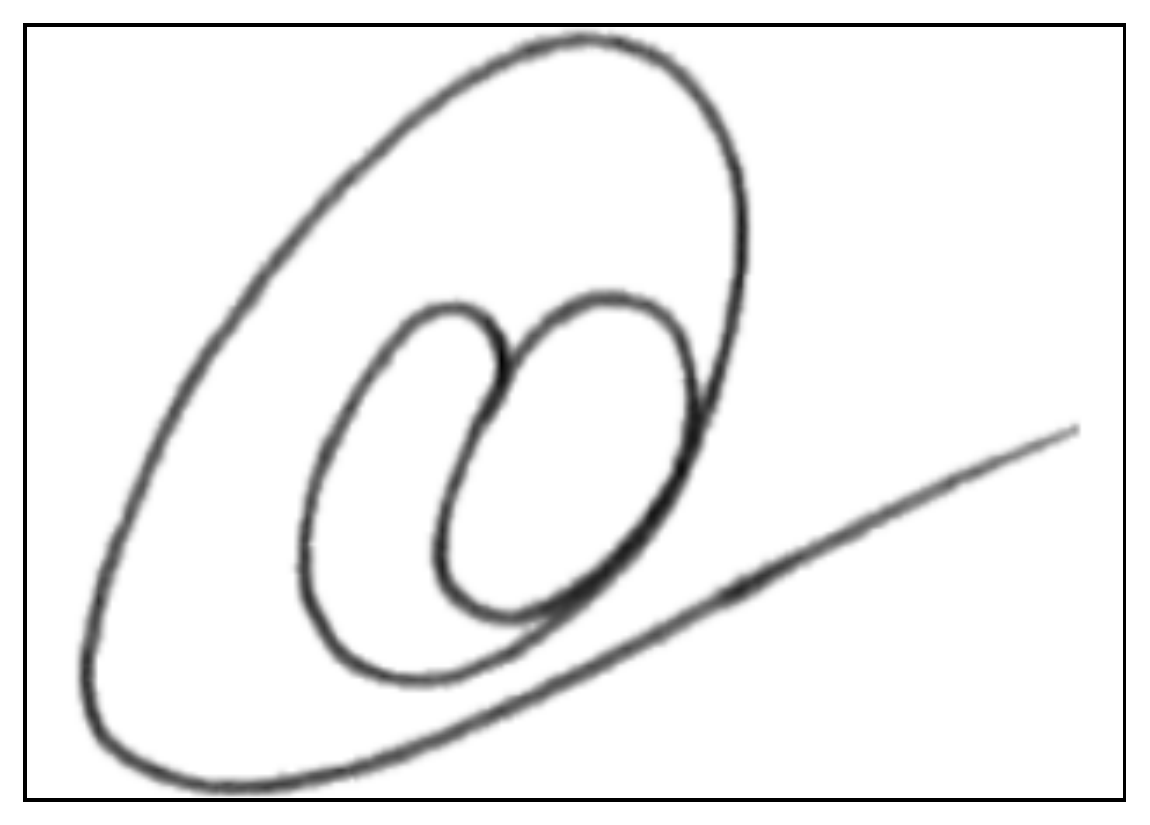} 
\qquad
 \includegraphics[scale=0.28]{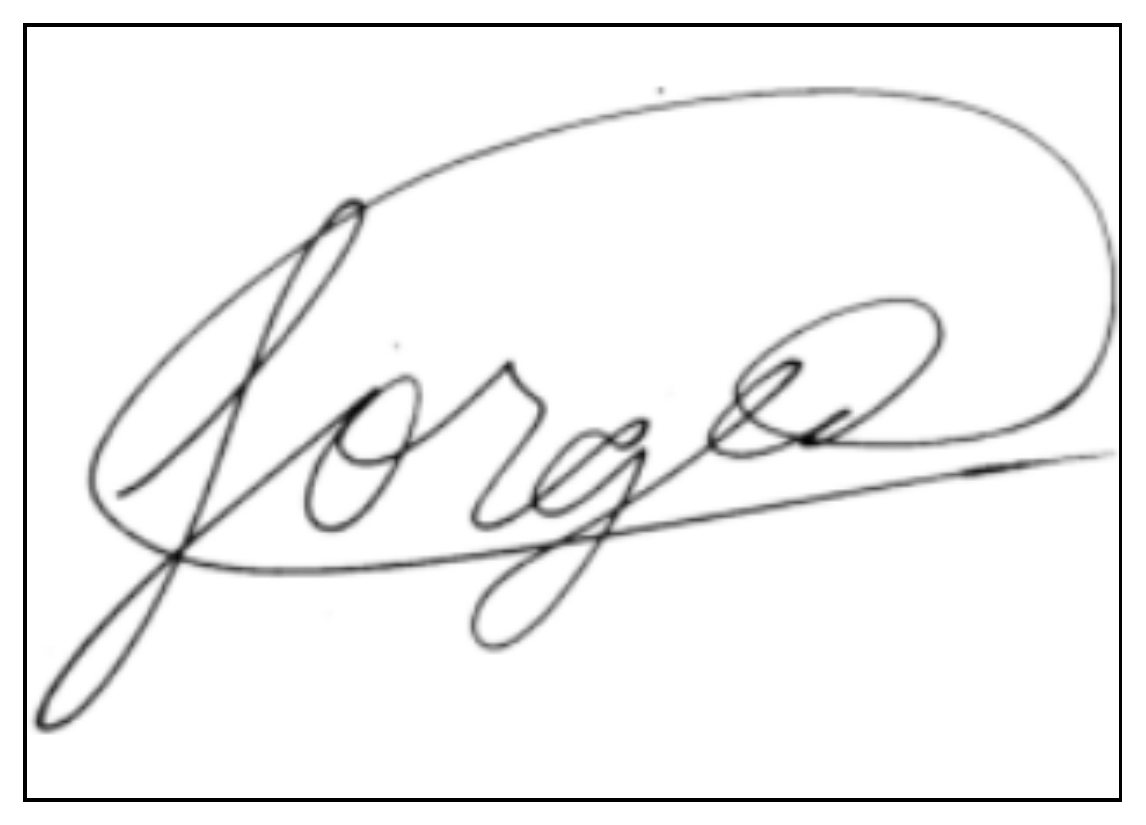}}
\qquad
\subfloat[Centering in a canvas and then resizing]{\includegraphics[scale=0.28]{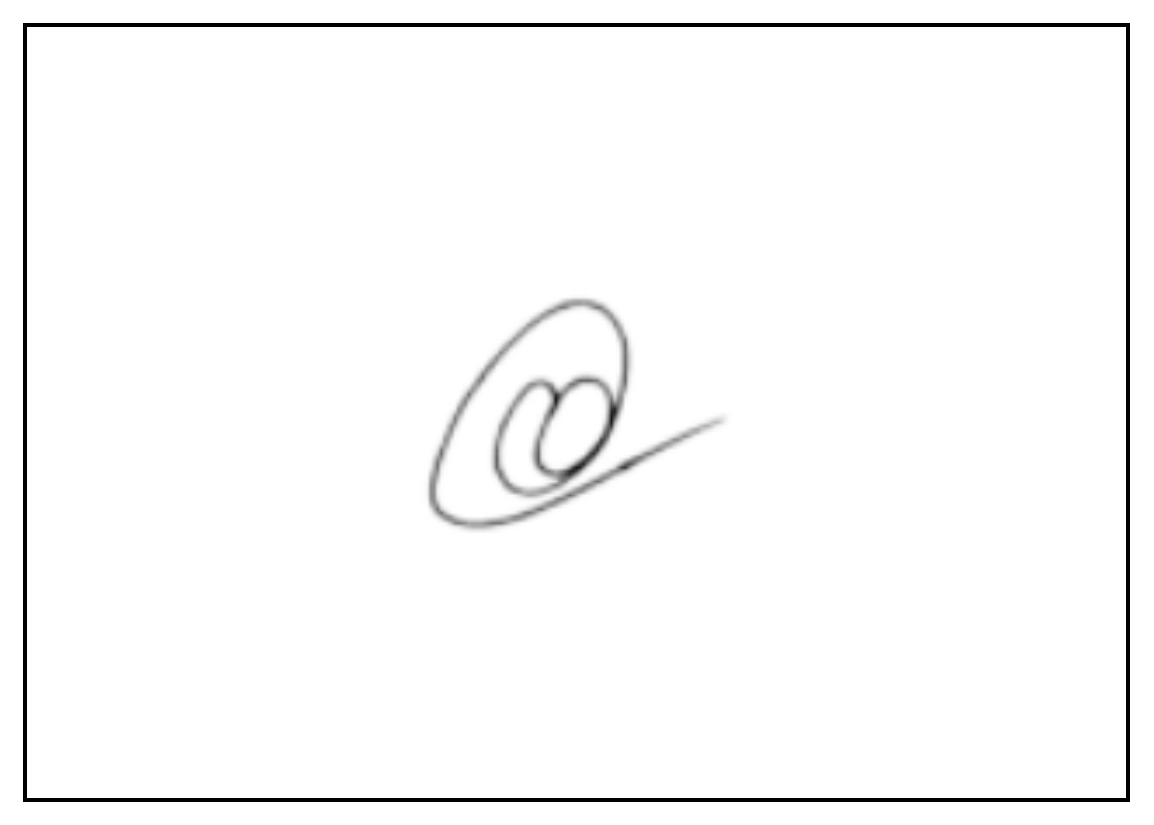}
\qquad
\includegraphics[scale=0.28]{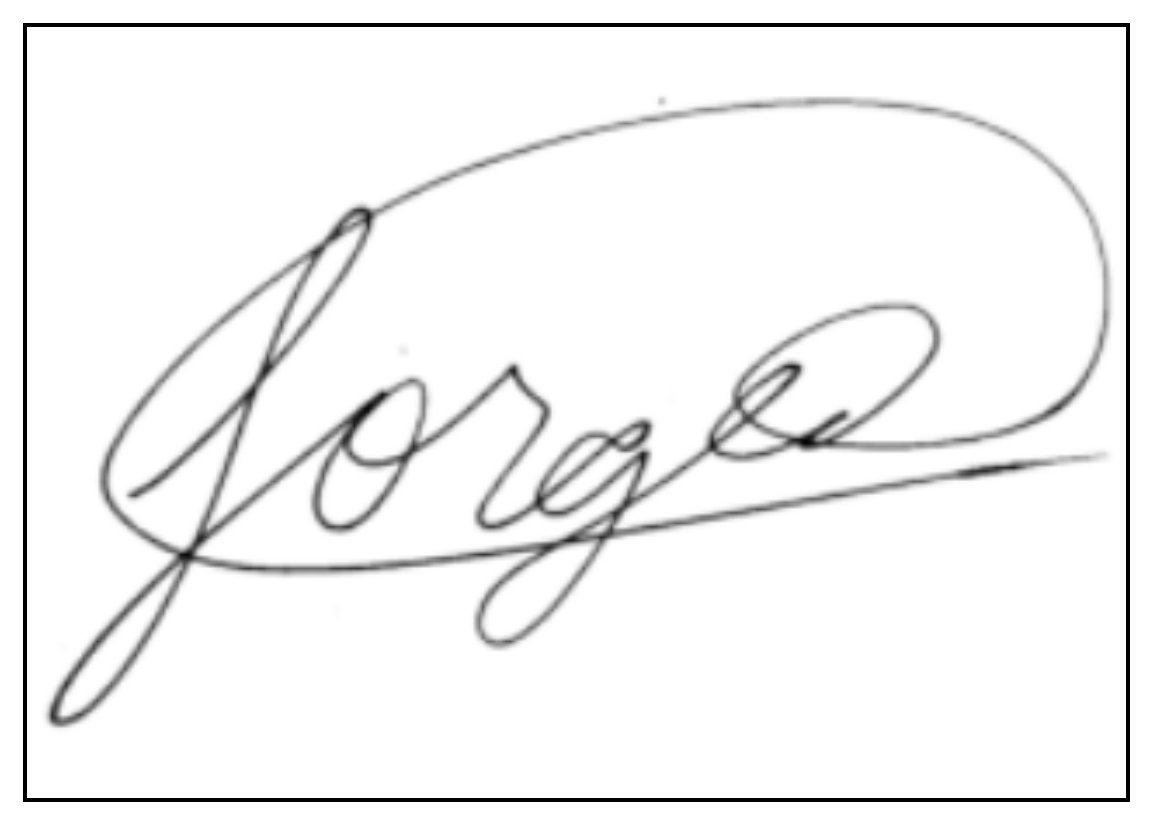}}
\caption{Two approaches for normalizing the signatures to a common size. The signature on the left is small ($176\times 229$ pixels) while the signature on the right is large ($484\times 819$ pixels). (a) directly resizing the signatures to the input size of the network ($170\times 242$); (b) centering the signatures in a canvas of a ``maximum size" ($600\times 850$) and then resizing to $170\times 242$ pixels. }
\label{fig:resizing}
\end{figure}

Although these methods present good performance, they also introduce some issues. Signatures from different users vary significantly in size, while a feature descriptor should provide a fixed-sized representation for classification. This is not a problem in many feature descriptions used for signature verification, that by design are able to accommodate signatures of different sizes. Neural networks, on the other hand, in general require fixed-sized inputs, and thus these methods require pre-processing the signatures such that they all have the same size. Most commonly, signatures are either a) resized to a common size or b) first centered in a blank image of a ``maximum signature size", and then resized. Figure \ref{fig:resizing} illustrates the problems with these approaches. In alternative (a), the width of the strokes become very different depending on the size of the original image, while in alternative (b) the width of strokes is not affected, but instead we may lose detail on small signatures, that would otherwise be preserved in the first alternative. Empirically, alternative (b) presented much better results \cite{hafemann_ijcnn_2016}, but it also creates the problem that now a ``maximum size" is defined, and if a new signature is larger than this size, it would need to be reduced (causing similar problems to (a) regarding the width of the strokes).

\begin{figure}
\centering
\subfloat[Genuine, 300 dpi]{
\includegraphics[scale=0.5]{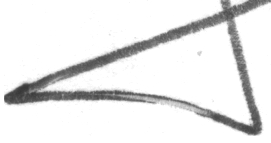}
}
\subfloat[Genuine, 100 dpi]{
\makebox[0.4\columnwidth]{%
\includegraphics[scale=1]{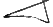}
}%
}
\qquad
\subfloat[Forgery, 300 dpi]{
\includegraphics[scale=0.5]{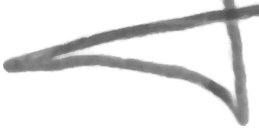}
}
\subfloat[Forgery, 100 dpi]{
\makebox[0.4\columnwidth]{%
\includegraphics[scale=1]{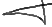}
}%
}
\caption{Detail of a genuine signature and a skilled forgery for user 244 in the GPDS dataset. At 300 dpi, we can notice the limp strokes of the skilled forgery, most likely due to slow hand movements while attempting to reproduce the overall shape of the genuine signature. At 100 dpi, information about line quality is mostly lost.}
\label{fig:linequality}
\end{figure}

Another problem in learning the representations from signature images is the selection of the resolution of the input images. The methods proposed in the literature use small images (e.g. $96 \times 192$ in \cite{rantzsch_signature_2016}, $170 \times 242$ in \cite{hafemann_icpr_2016}). For the signatures used in these papers, this is equivalently of using a resolution around 100 dpi. However, as illustrated in figure \ref{fig:linequality}, the distinction of genuine signatures and skilled forgeries often rely on the line quality of the strokes (in particular for slowly-traced forgeries, as noted in \cite{hafemann_icpr_2016}). This suggests that using higher resolutions may improve performance on this task.

In this paper, we propose learning a fixed-sized representation for signatures of variable size, by adapting the architecture of the neural network, using Spatial Pyramid Pooling (SPP) \cite{he_spatial_2014}, \cite{he_spatial_2015}. Our contributions are as follows: we define and evaluate different training protocols for networks with SPP applied to offline handwritten signatures. After training, signatures of any size can be fed to the network in order to obtain a fixed-sized representation. 
We also evaluate the impact of the image resolution on the classification accuracy, and the generalization of features learned in one dataset to other operating conditions (e.g. different acquisition protocols, signatures from people of different locations), by using transfer learning to other datasets. 

For feature learning, we used the problem formulation presented in \cite{hafemann_pr_2017}, where Writer-Independent features are learned using a subset of users, and subsequently used to train Writer-Dependent classifiers for another set of users. We also use the architecture defined in this work as baseline (SigNet). We adapt this architecture to learn fixed-sized representations (proposing different training protocols) and modifying the architectures to handle images of higher resolution.
We conducted experiments on four widely used signature verification datasets: GPDS, MCYT, CEDAR and the Brazilian PUC-PR dataset; and two synthetic datasets (Bengali and Devanagari scripts).
Using the proposed architecture, we obtain a similar performance compared to the state-of-the-art, while removing the constraint of having a fixed maximum signature size. We also note that using higher resolutions (300dpi) greatly improves performance when skilled forgeries (from a subset of users) is available for training. On the other hand, if only genuine signatures are used for feature learning, higher resolutions did not improve performance. We verify that the learned features generalize to different operating conditions (by testing them on other datasets), and that fine-tuning the representation for the different conditions further improves performance. We observed that the features learned on GPDS generalize better to other western signature datasets (MCYT, CEDAR and Brazilian PUC-PR) than to other types of scripts (Bengali and Devanagari), and that fine-tuning also largely addresses this problem.

\section{Related Work}

The problem of Offline Signature Verification is either formulated as Writer-Dependent, with one classification task defined for each user enrolled to the system, or as a Writer-Independent problem, where we consider a single problem, of comparing a questioned signature to a reference signature. 
In the literature, Writer-Dependent classification is most commonly used: for each user, a set of reference (genuine) signatures are used as positive samples, and a set of genuine signatures from other users (in this context called ``Random forgeries") are used as negative samples, and a binary classifier is trained. Alternatively, some authors propose using one-class classifiers for the Writer-Dependent formulation, using only genuine signatures from the user as positive samples (e.g. \cite{guerbai_effective_2015}). 
Writer-Independent classification, on the other hand, is often used by training a binary classifier on a dissimilarity space, where the inputs are the absolute difference of two feature vectors: $\textbf{x} = |\textbf{f}_1 - \textbf{f}_2|$, where $\textbf{f}_1$ and $\textbf{f}_2$ are feature vectors extracted from two signatures, and we consider a binary label: $y = 1$ if both signatures are from the same user, and $y = -1$ otherwise \cite{bertolini_reducing_2010,rivard_multi-feature_2013,eskander_hybrid_2013}.

After training the classifiers, we verify the performance of the system in distinguishing genuine signatures from forgeries. We adopt the following definitions of forgery, which are the most common in the Pattern Recognition community: ``Random Forgeries" are forgeries made without any knowledge of the user's genuine signature, where the forger uses his own signature instead. In the case of ``Simple forgeries", the forger has access to the person's name. In this case, the forgery may present more similarities to the genuine signature, in particular for users that sign with their full name, or part of it. Lastly, for ``Skilled Forgeries", the forger has access to the user's signature, and often practices imitating it. This result in forgeries that have higher resemblance to the genuine signature, and therefore are harder to detect. While discriminating Random and Simple forgeries are relatively simpler tasks (as reflected in lower error rates in the literature), discriminating genuine signatures and skilled forgeries remains a challenging task.

A critical aspect of designing signature verification systems is how to extract discriminant features from the signatures. A large part of the research efforts on this field addresses this question, by proposing new feature descriptors for the problem. These features range from simple descriptors such as the size of the signature and inclination \cite{nagel_computer_1977}, graphometric features \cite{justino_off-line_2000}, \cite{oliveira_graphology_2005}, texture-based \cite{vargas_off-line_2011, yilmaz_score_2016}, interest point-based \cite{pal_off-line_2012}, among others. Recent advancements in this field include using multiple classifiers trained with different representations \cite{yilmaz_score_2016}, using interval symbolic representations \cite{alaei_efficient_2017} and augmenting datasets by duplicating existing signatures or creating synthetic ones \cite{ferrer_static_2015, diaz_generation_2016, ferrer_static_2017}. More recently, methods for learning features from signature images have been proposed \cite{hafemann_ijcnn_2016, hafemann_pr_2017, rantzsch_signature_2016,zhang_multi-phase_2016}. Although these methods demonstrated improved performance, they introduced some issues, notably by requiring that all signature images have the same size, which is the problem addressed in this paper. We note that this problem is not present in many handcrafted feature descriptions used for signature verification: for instance Local Binary Patterns (LBP) \cite{ojala_multiresolution_2002} and Histogram of Oriented Gradients (HOG) \cite{dalal_histograms_2005} use histograms over the entire image, therefore resulting in feature vectors of the same size regardless of the input size; Extended Shadow Code (ESC) \cite{sabourin_extended-shadow-code_1993} divides the image in the same number of windows (adapting the size the windows), therefore also working with signatures of variable sizes.

The problem of requiring inputs of a fixed size for neural networks also affects other applications, such as object recognition. This problem is often handled by simply resizing and cropping images. While these are common operations for object recognition, we argue that they are less interesting for signature verification. In object recognition, the classification task considers objects at different scales. Therefore, resizing an image to fit a particular size is a reasonable action to take, since it is aligned with the invariance to scale that we expect from the classifiers (as long as the change in scale does not distort the image, such as scaling height and width by different factors). On the other hand, for signature verification we have control of how the signatures are acquired: all signatures are usually scanned at the same resolution, usually 300 or 600 dpi. Therefore, changes in scale, introduced by resizing the image, alter the signal is ways that would not otherwise be present. In this case, a better solution would not require resizing the signature images by different factors depending on their original size.

In the context of object recognition, He at al. proposed a solution for working with inputs of variable size, by using Spatial Pyramid Pooling \cite{he_spatial_2015}. However, the training procedure still requires fixed-sized images. In \cite{he_spatial_2015} the authors proposed using two image sizes, resizing each image to the these sizes (i.e. duplicating the dataset in two different scales). Learning is then conducted by alternating between the two sets in each training epoch. This is sub-optimal for signature images, since we would like to avoid resizing the images entirely. In this work we propose and test other training protocols for training networks with SPP on signature data.

\section{Proposed Method}

In this work we consider the two-stage approach described in \cite{hafemann_ijcnn_2016} and \cite{hafemann_pr_2017}, where we train Writer-Dependent classifiers on a set of users, using a feature representation learned on another set of users. We note, however, that the methods described in this paper can be used for other feature learning strategies, such as the ones used in \cite{rantzsch_signature_2016, zhang_multi-phase_2016}.

We consider two disjoint sets of users: a development set $\mathcal{D}$, where we learn feature representations, and an exploitation set $\mathcal{E}$ that consists of the users ``enrolled to the system", for whom we train Writer-Dependent classifiers. The first phase consists in learning a function $\phi(X)$, using the data from $\mathcal{D}$, that takes a signature $X$ as input, and returns a fixed-sized feature vector. In the second phase, we use this learned function to ``extract features" for the signatures in $\mathcal{E}$, and train a binary classifier for each user. While we could use all users for learning the representations, this separation in two sets allows us to estimate the generalization performance of using this learned representation for new users. This is important since the set of users in a system is not fixed - new users may enroll at any time, and in this formulation we simply use the learned function $\phi(X)$ to obtain a representation for the signatures of this new user, and train a binary classifier.

In order to handle signatures of different sizes, we change both the feature learning process, as well as the process to obtain representations for new signatures using the learned network.

\subsection{Network architecture and objective function}
\label{sec:arch_loss}

\begin{figure*}
\includegraphics[width=\textwidth]{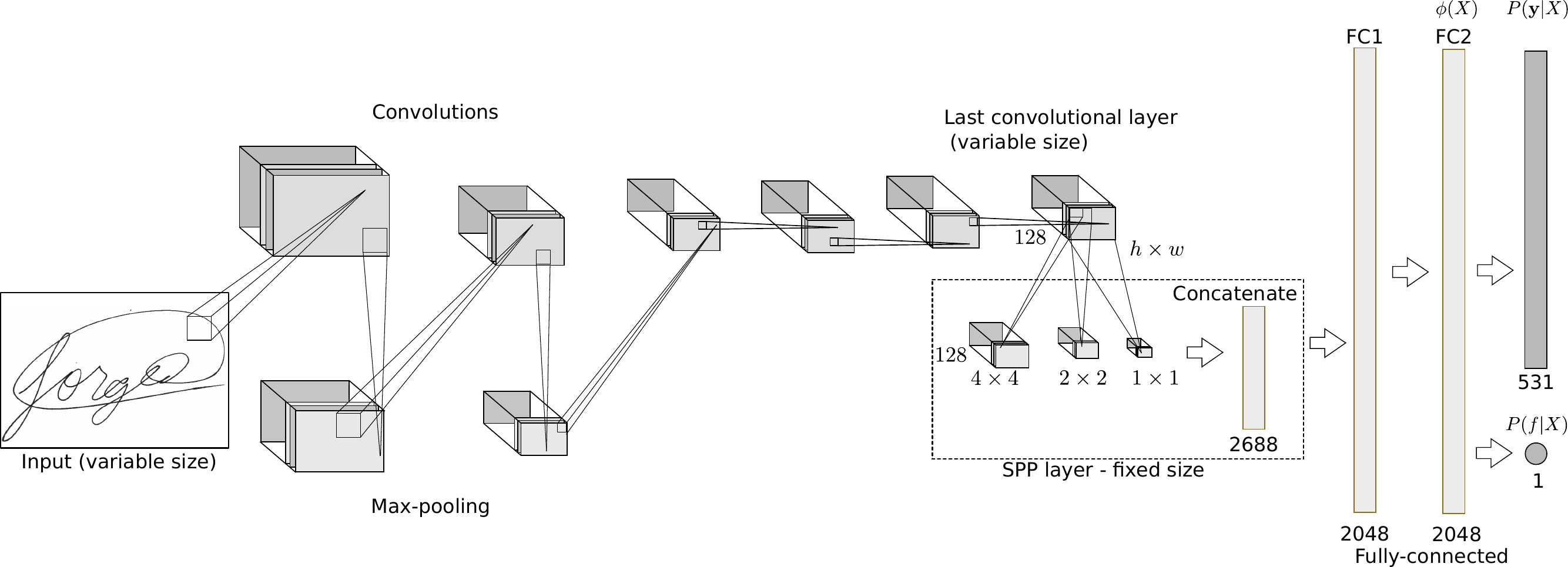}
\caption{One of the CNN architectures used in this work. The input signature (of variable size) is transformed in a sequence of convolution and max-pooling operations. The last convolutional layer results in 128 maps of size $h \times w$ (the actual size varies according to the signature size). The Spatial Pyramid Pooling layer (SPP) is then used to obtain a fixed-sized representation, by adapting the size of pooling regions, to obtain pooled results in three sizes: $4\times 4$, $2\times 2$ and $1\times 1$. These are concatenated in a single vector of size $21 \times 128 = 2688$ units, which is then used as input to the fully-connected layers. During training, the network outputs $P(\textbf{y}|X)$ (and, if forgeries are used during training, also $P(f|X)$), and the network is trained to minimize the cross-entropy with respect to the training dataset. For obtaining representations for new signatures, we perform forward propagation until the last layer before softmax, obtaining $\phi(X)$, a vector of 2048 dimensions, regardless of the signature size.}
\label{fig:cnn_architecture}
\end{figure*}

The definition of a Convolutional Neural Network architecture usually specifies the input size of the images for training and testing. However, as noted in \cite{he_spatial_2015}, this constraint is not caused by the usage of convolution and pooling layers, but rather by the usage of fully-connected layers at the end of network architectures. The reason is that the convolution and pooling operations are well defined for inputs of variable sizes, simply resulting in an output of a larger size. This presents a problem between the last pooling layer and the first fully-connected layer of the network (layer FC1 in figure \ref{fig:cnn_architecture}): the last pooling layer is ``flattened" to  a vector of dimensionality K (e.g. a pooling output of size $32 \times 3\times 2$ becomes a vector of $K=192$ elements), and the fully-connected layer uses a weight matrix of size $K \times M$, where M is the output size of the fully-connected layer. If we use the network to process an input $\hat{X}$ of a different size, the output of the last pooling layer will have a different size. Flattening the representation results in a vector of dimensionality $\hat{K} \neq K$, and therefore the vector-matrix product in the fully-connected layer will not be defined.

The central idea of Spatial Pyramid Pooling (SPP) \cite{he_spatial_2015} is to obtain a fixed-size representation for variable-sized input images. This is done by adapting the size of the pooling region (and strides) for each image size, such that the output of the last pooling operation has a fixed size, and therefore can be used as input to fully-connected layers. In SPP, a set of fixed-sized outputs is chosen, and the result of them is concatenated. This is illustrated in the ``SPP Layer" box in figure \ref{fig:cnn_architecture}: we consider pooling with output sizes  $1\times 1$, $2 \times 2$ and $4 \times 4$, which has a total of $1 + 4 + 16 = 21$ outputs for each channel. Each image would therefore output a fixed representation of size $21 \times C$, where $C$ is the number of feature maps/channels in the last convolutional layer.

Figure \ref{fig:cnn_architecture} illustrates a CNN architecture used in this work. The network contains a series of convolutions and max-pooling operations, with a Spatial Pyramid Pooling layer between the last convolutional layer and the first fully-connected layer. This layer outputs a fixed-sized output regardless of the size of the input signature. 

We consider two application scenarios, as in \cite{hafemann_pr_2017}: one in which we have only genuine signatures available for training, and one in which we also have access to skilled forgeries for a subset of users. In the first scenario, we consider a training objective of distinguishing between different users in the development set: the network outputs $P(\textbf{y}|X)$: the probability of the signature belonging to one of the users in $\mathcal{D}$. Therefore, the network learns to identify the users that produced the signatures in $\mathcal{D}$. In the second scenario, we would like to leverage the information of forgeries in the feature learning process, and we use the multi-task approach defined in \cite{hafemann_pr_2017}. In this formulation, the network also predicts whether or not the signature is a forgery: $P(f|X)$. We simultaneously train the network to optimize both objectives (distinguish between different users, and between genuine signatures and skilled forgeries), by using the loss function defined in equation \ref{eq:loss_forg}:

\begin{equation}
\begin{split}
L &= (1- f_i) (1-\lambda) L_c + \lambda L_f \\
&= -(1- f_i) (1-\lambda) \sum_j{y_{ij} \log{P(y_{j} | X_i})} + \\
& \lambda \big(-f_i \log(P(f | X_i)) - (1 - f_i) \log (1 - P(f | X_i))\big)
\end{split}
\label{eq:loss_forg}
\end{equation}

\noindent Where $\lambda$ is a hyperparameter that trades-off between the two objectives, $X_i$ is the signature, $\textbf{y}_i$ is the actual user of signature ($y_{ij} = 1$ if the signature $i$ belongs to user $j$), and $f_i$ indicates whether or not the signature $i$ is a forgery. $L_c$ and $L_f$ indicate the loss functions for user classification and forgery classification, respectively, which are expanded in the second and third lines. We refer the reader to \cite{hafemann_pr_2017} for more details on this formulation.

\begin{table*}
%\ra{1.3}
\centering
\caption{CNN architectures used in this paper}
\label{tbl:architectures}
\begin{tabular}{cc|cc|cc}
\toprule
SigNet&SigNet-SPP&SigNet-300dpi&SigNet-SPP-300dpi&SigNet-600dpi&SigNet-SPP-600dpi\\ \midrule
conv11-96-s4-p0&conv11-96-s4-p0&conv11-32-s3-p5 &conv11-32-s3-p5 &conv11-32-s4-p5 &conv11-32-s4-p5 \\ 
pool3-s2-p0&pool3-s2-p0&pool3-s2-p0&pool3-s2-p0&pool3-s3-p0 &pool3-s3-p0 \\ 
conv5-256-p2 &conv5-256-p2 &conv5-64-p2 &conv5-64-p2 &conv5-64-p2 &conv5-64-p2 \\ 
pool3-s2-p0  &pool3-s2-p0  &pool3-s3-p0  &pool3-s3-p0  &pool3-s2-p0 &pool3-s2-p0 \\ 
conv3-384-p1&conv3-384-p1&conv3-128-p1&conv3-128-p1&conv3-128-p1 &conv3-128-p1 \\ 
conv3-384-p1&conv3-384-p1&conv3-128-p1   &conv3-128-p1   &conv3-128-p1 &conv3-128-p1 \\ 
&&pool3-s2-p0 &pool3-s2-p0 &pool2-s2-p0 &pool2-s2-p0 \\ 
conv3-256-p1&\textbf{conv3-180-p1}&conv3-128-p1   &conv3-128-p1   &conv3-128-p1 &conv3-128-p1 \\ 
pool3-s2-p0&\textbf{spp-4-2-1} &pool3-s3-p0  &\textbf{spp-4-2-1} &pool4-s4-p0&\textbf{spp-4-2-1} \\ 
\midrule
\multicolumn{6}{c}{FC1-2048} \\ 
\multicolumn{6}{c}{FC2-2048} \\ 
\multicolumn{6}{c}{FC-M + softmax ; FC-1 + sigmoid} \\ 
\bottomrule
\end{tabular}
\end{table*}

Table \ref{tbl:architectures} lists the CNN architectures used in this paper. We consider a total of six architectures, considering three different resolutions (around 100 dpi, 300 dpi and 600 dpi), and with or without Spatial Pyramid Pooling. 
Each line in the table represents a layer of the CNN. For convolutional layers, we specify the size and number of feature maps (filters), the stride and the padding. For instance, conv11-32-s4-p5 refers to a convolutional layer with 32 filters of size $11 \times 11$, with stride $s=4$ and padding $p=5$. For pooling operations, we inform the pool size, the stride and padding. When not specified, we use stride $s=1$. After each learnable layer (with the exception of the output layers) use a Batch Normalization layer \cite{ioffe2015batch}. The SigNet architecture was defined in previous work \cite{hafemann_pr_2017}, while the other architectures are adapted versions to handle larger images. For higher resolutions, we notice that images are very large (e.g. 780x1095 pixels for a 600 dpi signature). To handle these larger images, we used a smaller number of feature maps, and a more rapid reduction in size across the layers, by using a more aggressive pooling. For each of the three resolutions, we consider both a version with SPP (that accepts inputs of any size), and without SPP (that accepts inputs of a fixed size). The two versions have the same overall structure, but diverge on the last pooling layer. The network SigNet-SPP has another difference (lower number of convolutional maps in the last convolutional layer) to keep the number of parameters between the SPP and non-SPP version similar. In the table, the differences between the non-SPP and SPP versions are highlighted in bold.
In all architectures, the last layer outputs $M$ neurons, which  estimate $P(\textbf{y} | X)$, the probability of a signature $X$ belonging to a particular user in $\mathcal{D}$. For the experiments using forgeries during feature learning, the network also outputs a single neuron that predicts $P(f|X)$, the probability that the signature is a forgery. We report experiments with both scenarios (with and without forgeries for feature learning). In the cases were forgeries are used, we append a suffix \textbf{-F} to the architecture name. For example,  \emph{SigNet-300dpi-F} refers to using the architecture SigNet-300dpi  using both genuine signatures and skilled forgeries for training, while \emph{SigNet-300dpi} refers to using the same architecture, but trained with only genuine signatures.

The Spatial Pyramid Pooling layer was implemented as in \cite{he_spatial_2015}: we use pooling regions of sizes $4 \times 4$, $2 \times 2$ and $1 \times 1$, resulting in a total of 21 outputs for each feature map. The pooling region and strides are dynamically determined for each input size. Let $h$ and $w$ be the output size of the last convolutional layer. For the pyramid level of size $n \times n$, the pooling region of the unit $(j,i)$ is defined as rows between: $\big\lfloor \frac{j-1}{n}h \big\rfloor$ and $\big\lceil \frac{j}{n}h \big\rceil$ and columns between  $\big\lfloor \frac{i-1}{n}w \big\rfloor$ and $\big\lceil \frac{i}{n}w \big\rceil$, where $\lfloor . \rfloor$ and $\lceil . \rceil$ denote the floor and ceiling operations. Similarly to max-pooling, we take the max of this pooling region, that is, the output of unit $(j,i)$ is the maximum value of the pooling region defined above. The implementation of this layer has been made publicly available in the Lasagne library \footnote{\url{https://github.com/Lasagne/Lasagne/}}.

\subsection{Training protocol}
\label{sec:training_protocol}

The neural networks are initialized with random weights following \cite{glorot_understanding_2010}, and training is performed with stochastic gradient descent to minimize the loss function defined in section \ref{sec:arch_loss}. We use mini-batches of data (which is required in order to use Batch Normalization, and also speeds up training), and we consider different protocols for generating the mini-batches, as described below.

The networks without SPP require a fixed input size for all images. In this case, we pre-process all signatures by centering them in a canvas of a ``maximum size", and then resizing to the desired input size for the network.

When using Spatial Pyramid Pooling, we can process signatures of any size, but during training we need to design a protocol that provides batches of signatures having the exact same size. We consider two alternatives:

\begin{enumerate}
\item \textbf{Fixed size}: Using a single ``maximum signature size" (as in the training for networks without SPP);
\item \textbf{Multiple sizes}: Defining multiple canvas sizes, and centering each signature on the smallest canvas that fit the signature.
\end{enumerate}

In the first case, all the images on the training set  have the same size, and we simply process the images in mini-batches in a random order.

For the second alternative, we define different image sizes based on statistics of the development set (the set of signatures used to train the CNN). Consider the following definitions:
\begin{itemize}
\item $\mu_h$, $\sigma_h$, $\text{max}_h$: height of the signatures in the development set (mean, standard deviation and maximum, respectively);
\item $\mu_w$, $\sigma_w$, $\text{max}_w$: width of the signatures in the development set (mean, standard deviation and maximum, respectively).
\end{itemize}

We divide the dataset into 5 different parts, as follows:
\begin{enumerate}
\item Images larger than 3 standard deviations are considered ``outliers". In particular, images taller than $\tau_h = \mu_h + 3 \sigma_h$ or wider than $\tau_w = \mu_w + 3 \sigma_w$ are all assigned to the largest canvas, of size ($\text{max}_h \times \text{max}_w$);
\item The remaining signatures are split in four groups, by using the medians of the height and width. Given the medians $\tilde{H}$ and $\tilde{W}$ for the height and width, respectively, we consider the following canvas sizes: ($\tilde{H} \times \tilde{W}$), ($\tilde{H} \times \tau_w$), ($\tau_h \times \tilde{W}$), ($\tau_h \times \tau_w$).
\end{enumerate}

Each signature is centered (not resized) in the smallest canvas size that fits the signature. Therefore, this creates a total of 5 datasets, one for each canvas size. 

During training, we create an iterator for each of the 5 datasets: each iterator returns batches of signatures of the same size (within the batch). We then train the model by taking batches of different image sizes, alternating the sizes after each mini-batch (contrary to \cite{he_spatial_2015} that alternated after an entire epoch).  This procedure is detailed in Algorithm \ref{alg:train}. In this algorithm, the \emph{train} function is a single step of Stochastic Gradient Descent, with Nesterov Momentum.

\SetKwFunction{FRtrain}{TrainWithMultipleSizes}%
\begin{algorithm}
\SetKwProg{Fn}{function}{\string:}{}
\Fn(\tcc*[h]{train the network for one epoch}){\FRtrain{$S$, iterators}}{
\KwData{$S$: set of image sizes; \textbf{iterators}: list of data iterators, for each image size}

active $\leftarrow$ S \;
\While{active $\neq \emptyset$}{
  \For{$s \in active$}{
       \If{iterator[s].has\_next\_batch()}{
   	     	 mini\_batch $\leftarrow$ iterator[s].next\_batch() \;
	         train(mini\_batch) \;
       }
       \Else{
           active $\leftarrow$ active $\setminus$ s
       }
  }
}
}
\caption{Training algorithm for ``multiple sizes", for one epoch.}
\label{alg:train}
\end{algorithm}

We trained the networks for 60 epochs, using mini-batches of size $32$, L2 penalty with weight decay set to $10^{-4}$ and momentum factor of $0.9$. Training started with a learning rate of $10^{-3}$, which was decreased twice (at epochs 20 and 40) by dividing it by 10 each time.

\subsubsection{Data augmentation}

In previous work \cite{hafemann_icpr_2016} we performed data augmentation by performing random crops of the input images. We adopt the same protocol for the ``Fixed size" training.  However, in the ``Multiple sizes" protocol defined above, where we use smaller canvas sizes, cropping the images could result in cropping part of the signature, not only the background. Instead, we use the opposite strategy, of enlarging the signature images, by padding the signatures with the background color. We use this strategy to avoid losing part of the signal due to cropping. For example, consider an image of size 300x300, and a padding of size 20x20: we pad the image so that it has size 320x320, positioning the original image to randomly start between 0 and 20 pixels in height and width (i.e. not necessarily in the center of this new image). For even greater variability, we consider a maximum value of padding, and in each mini-batch we randomly select the padding between 0 and this maximum value.

\subsection{Fine-tuning representations}
\label{sec:finetuning}

When considering the generalization of the learned features to new operating conditions (e.g. new acquisition protocol), it is possible to fine-tune the representations to the new conditions. In order to evaluate the impact of fine-tuning the representations, we consider a network trained in one dataset as a starting point, and subsequently train it for users of another dataset. 

Similarly to previous work on transferring representations \cite{oquab_learning_2014, chatfield_return_2014}, we perform the following steps for fine-tuning representation to a new dataset:

\begin{itemize}
\item Duplicate the network that was trained in the first dataset;
\item Remove the last layer (that correspond to $P(\textbf{y}|X)$ for the users in the first dataset);
\item Add a new softmax layer, with $M_2$ units, corresponding to $P(\textbf{y}|X)$, the probability of a signature image belonging to one of the $M_2$ users of the second dataset;
\item Train the network on the second dataset with a reduced learning rate ($5 \times 10^{-4}$).
\end{itemize}

The training procedure during finetuning is similar to the training algorithm used for learning the features in the first dataset. The exception is for the SPP models trained with a ``Fixed size". In this case, we consider two distinct sizes: the original maximum signature size from the first dataset, and the maximum signature size from the target dataset.

Since different datasets have different acquisition protocols (e.g. type of writing instrument, instructions for the forgers, and the resolution of scanned images), we expect that fine-tuning the representations to a set of users from the same domain should improve performance.

\subsection{Training WD classifiers}
\label{sec:training_wd}

After we learn the CNN in one set of users, we use it to obtain representations for signatures of users in the exploitation set $\mathcal{E}$, and train Writer-Dependent classifiers. The procedure to obtain the representations vary slightly depending on the training method:

\begin{itemize}
\item Networks without SPP: The signatures from $\mathcal{E}$ are centered in a canvas of maximum size ($H_\text{max} \times W_\text{max}$). During transfer learning, we consider the maximum size of the target dataset, and resize all images to the size of the original dataset;
\item SPP trained with ``fixed size": The signatures from $\mathcal{E}$ are centered in a canvas of size ($H_\text{max} \times W_\text{max}$). During transfer learning, signatures that are larger than this canvas are processed in their original size;
\item SPP trained with ``multiple sizes": The signatures are processed in their original size.
\end{itemize}

The differences among the three training alternatives are summarized in table \ref{tbl:differences_sppnonspp}.

For each user in the set $\mathcal{E}$, we build a dataset consisted of $r$ genuine signatures from the user as positive samples, and genuine signatures from other users as negative samples. We then train a binary classifier for the user: a linear SVM or an SVM with the RBF kernel. We usually have many more negative than positive samples for training, since we only have a few genuine signatures for the user, while we can use samples from many users as negative samples. For this reason, we correct this skew by giving more weight to the positive samples, as described in \cite{hafemann_pr_2017}. After the classifiers are trained, we measure their capability of classifying genuine signatures are different types of forgery.

\begin{table}
\caption{Summary of differences between the training/testing protocols}
\label{tbl:differences_sppnonspp}
\resizebox{\columnwidth}{!}{%
\begin{tabular}{p{76px}p{70px}p{70px}p{70px}}
\toprule
&Without SPP&SPP (training with fixed size)&SPP (training with multiple sizes) \\ \midrule
Training images&Centered in a fixed size&Centered in a fixed size&Consider 5 different sizes \\ \cmidrule{2-4}
%Data augmentation&Crop images&Crop images&Enlarge (pad) images \\ \cmidrule{2-4}
CNN architecture&Use pooling with fixed pooling size&Use SPP (variable pooling size, fixed output)&Use SPP (variable pooling size, fixed output) \\ \cmidrule{2-4}
Generalization (extracting features)&Center images in a fixed size. Larger images are resized&Center images in a fixed size. Larger images are processed in their original size&All images are processed in their original size \\ \cmidrule{2-4}
Finetuning & Center images in the maximum size of the target dataset. All images are resized to the maximum size of the source dataset & Center images in two canvases: maximum size of the target dataset, and maximum size of the source dataset & Consider 5 different sizes (defined in the target dataset) \\
\bottomrule
\end{tabular}
}%
\end{table}

\section{Experimental Protocol}

We conducted experiments on four offline handwritten signature datasets: GPDS-960 \cite{vargas_off-line_2007}, MCYT-75 \cite{ortega-garcia_mcyt_2003}, CEDAR \cite{kalera_offline_2004} and Brazilian PUC-PR \cite{freitas_bases_2000}; and two synthetic datasets, for Bengali and Devanagari scripts \cite{ferrer_static_2017}. We used a subset of the GPDS-960 dataset for learning feature representations, using the different architectures and training methods described in this article. We then evaluate the performance of Writer-Dependent classifiers trained with these feature representations, on a disjoint subset of GPDS, as well as the other datasets.

In order to allow comparison with previous work, we used the development set $\mathcal{D}$ as the last 531 users of GPDS (users 350-881) for training the CNNs. For the training protocol using ``multiple sizes", we followed the procedure detailed in section \ref{sec:training_protocol} to process the development dataset into 5 different canvas sizes. For instance, at 600 dpi we used canvases of size $338 \times 684$, $338 \times  1183$, $619 \times  684$, $619 \times  1183$ and $778 \times  1212$. The networks were then trained as defined in section \ref{sec:training_protocol}.

The images were pre-processed to remove noise, by applying OTSU's algorithm to find the threshold between background and foreground. The background pixels were set to white, leaving the signature pixels in grayscale. The images were then inverted by subtracting them from the maximum pixel intensity: $I_P(x,y) = 255 - I(x,y)$. In the resulting images the background is therefore zero-valued. The OTSU algorithm was not applied to the two synthetic datasets, since they do not contain any noise.

In the literature, slightly different protocols are used for each dataset, in particular regarding how many reference signatures are used for training, and which signatures are used as negative samples. We use the following protocols:
In the GPDS dataset, we trained Writer-Dependent classifiers for the first 300 users (to compare to results using the GPDS-300 dataset), using $r=12$ reference signatures as positive samples. We used $12$ signatures from each user in the development $\mathcal{D}$ as negative samples ($12 \times 531 = 6372$ signatures). This protocol is similar to the Brazilian dataset, where we have a separate development set $\mathcal{D}$. We train classifiers for the first 60 users using $r=15$, and 15 signatures from each of the remaining 108 users as negative samples ($15 \times 108 = 1620$ signatures).
In the MCYT and Cedar datasets, we used $r=10$ and $r=12$, respectively, and the same number of signatures from each other user in the exploitation set $\mathcal{E}$ as negative samples. In all cases, we trained a binary SVM, with an RBF kernel. We used the same hyperparameters as previous research \cite{hafemann_icpr_2016}: $C=1$ and $\gamma = 2^{-11}$, that were selected using a subset of the GPDS validation set. In this paper we did not explore optimizing these hyperparameters for each dataset (or even each user), but rather keep the same set of parameters for comparison with previous work.

For the experiments generalizing to different conditions (datasets), we considered two scenarios: using the CNN trained on GPDS to extract features without any changes, and fine-tuning the representation on these datasets. In these experiments, we used the network trained on GPDS images of the same resolution of the datasets (300 dpi for Cedar and Brazilian, 600 dpi for other datasets).
% In these experiments we also did not consider forgeries for learning the features

In order to assess the generalization performance of the fine-tuned representations, we conducted cross-validation experiments as follows:
\begin{enumerate}
\item The dataset in randomly split in two sets of users (50\%/50\%). Following the same terminology as before, we can consider them to be a development set $\mathcal{D}$ and exploitation set $\mathcal{E}$;
\item We fine-tune the CNN (originally trained on GPDS) for the development set $\mathcal{D}$;
\item We use the fine-tuned CNN to extract features for the exploitation set $\mathcal{E}$ and train WD classifiers.
\end{enumerate}

This protocol allows for an unbiased estimation of the performance on new users, whose signatures match the same operating characteristics of the dataset. We performed cross-validation running the steps above 10 times, each time randomly splitting the dataset in half, fine-tuning the CNN and training WD classifiers. For each fine-tuned CNN, we performed 10 runs on the WD classifier training with different signatures used for training/testing. Therefore, we fine-tuned a total of 10 CNNs for each dataset, and trained a total of 100 WD classifiers for each user in each dataset, and for each architecture. We then report the mean and standard deviation across these 100 runs. 

We evaluate the generalization of the learned representations to different scripts by training WD classifiers on synthetic signatures for two indian scripts: Bengali and Devanagari \cite{ferrer_static_2017}. For these datasets, since skilled forgeries are not available (the generation procedure is only defined for genuine signatures in \cite{ferrer_static_2017}) we evaluate the performance of the system on random forgeries. To allow for comparison with previous work, we train the WD classifiers with $r=5$ genuine signatures as positive samples. We also evaluated the errors on random forgeries on the other four datasets, which allows us to verify the generalization performance to other western scripts (on MCYT, CEDAR and Brazilian PUC-PR) and for other types of scripts (Bengali and Devanagari).

We evaluate the performance primarily using the Equal Error Rate (EER): which is the error when False Acceptance (misclassifying a forgery as being genuine) is equal to False Rejection (misclassifying a genuine as being a forgery). We considered two forms of calculating the EER: EER\textsubscript{user thresholds}: using user-specific decision thresholds; and EER\textsubscript{global threshold}: using a global decision threshold. For most experiments, we report the Equal Error Rates using only skilled forgeries. In the experiment where we compare the generalization to different scripts, we report the Equal Error Rates calculated with random forgeries.

For the Brazilian PUC-PR dataset, we used the same metrics as previous research in this dataset, and also report the individual errors (False Rejection Rate and False Acceptance Rate for different types of forgery) and the Average error rate, calculate as $\text{AER} = (\text{FRR} + \text{FAR\textsubscript{random}} + \text{FAR\textsubscript{simple}} + \text{FAR\textsubscript{skilled}}) / 4$. We also reported the average error rate considering only genuine signatures and skilled forgeries: $\text{AER\textsubscript{genuine + skilled}} = (\text{FRR} + \text{FAR\textsubscript{skilled}}) / 2$. 

For the comparison between different training types, and to measure the impact of finetuning, we use t-tests to compare the classifiers (using the EER\textsubscript{user thresholds} metric). We considered results significantly different for $p < 0.01$.

\section{Results}

We first present our analysis on using different image resolutions, followed by the analysis of the methods trained with SPP for handling signatures of variable size, and a comparison with the state-of-the-art.

\begin{figure}
\centering
\includegraphics[width=\columnwidth]{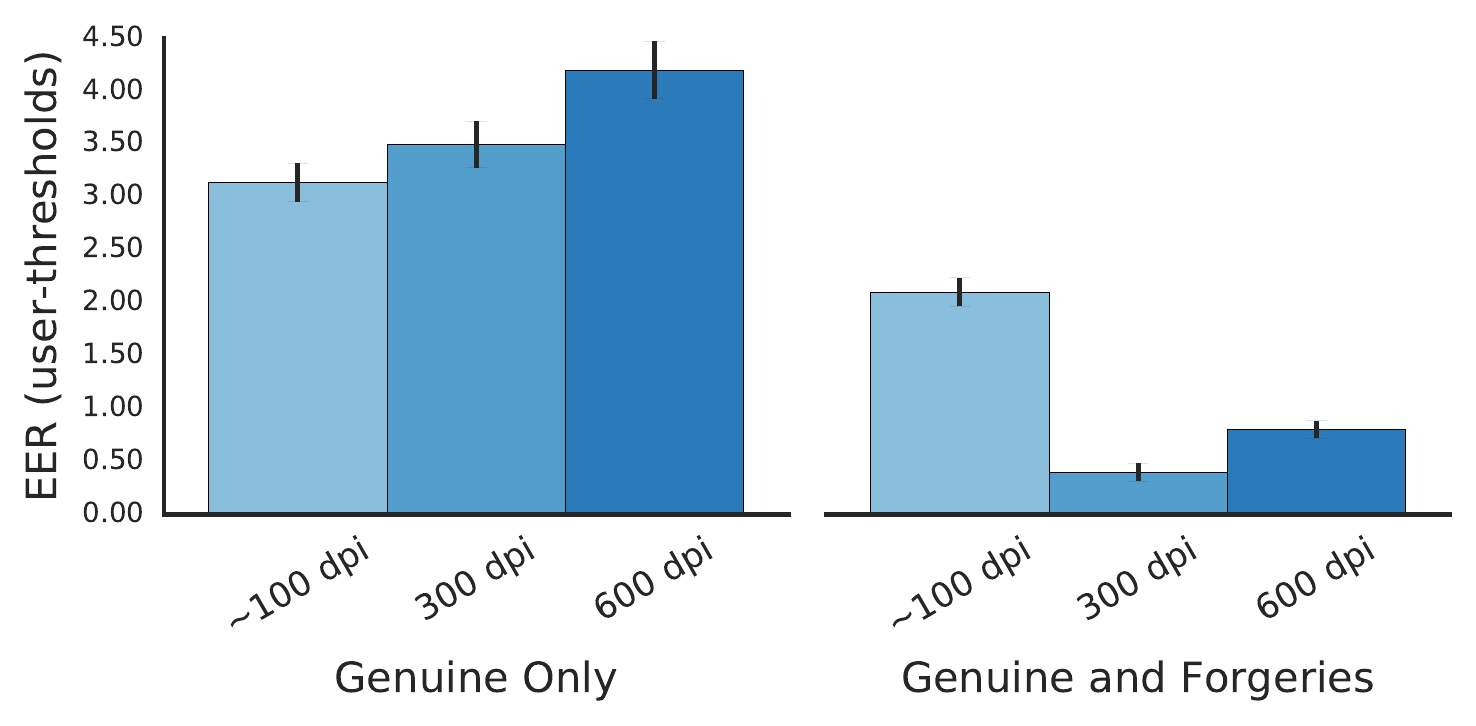}
\caption{Impact of the image resolution on system performance: EER of Writer-Dependent classifiers trained on GPDS-300, with representations learned in $\mathcal{D}$ at different resolutions. Left: Using only genuine signatures for feature learning; Right: Using genuine signatures and skilled forgeries for feature learning. Error bars indicate one standard deviation of the mean error (across 10 replications)}
\label{fig:varying_resolution}
\end{figure}

The results with varying the image resolution are summarized in figure \ref{fig:varying_resolution}. This figure shows the classification performance (EER) of Writer-Dependent classifiers trained on the GPDS-300 dataset, as we increase the resolution of the images. For these experiments, we consider the models trained without SPP, and consider two training scenarios: when we only use genuine signatures, and when skilled forgeries from a subset of users is used for feature learning (note that for training the WD classifiers, no skilled forgeries are used). The objective of this experiment is to verify the hypothesis that higher image resolutions are required to discriminate skilled forgeries. We notice an interesting trend in this figure: when using both genuine signatures and skilled forgeries, increasing the resolution greatly improves performance, reducing errors from 2.10\% using 100 dpi to 0.4\% using 300 dpi. On the other hand, increasing resolution did not improve performance when only genuine signatures are used for feature learning.  We argued in the introduction (in figure \ref{fig:linequality}) that low resolutions lose information about the line quality. These results suggest that, although fine details are present in higher resolution images, they are not taken into account when only genuine signatures are used for training the CNN. In other words, since the network does not have access to any skilled forgery, it does not learn features that discriminate line quality. Therefore, when only genuine signatures are available for training, low resolutions (100 dpi) are sufficient, but if forgeries from a subset of users are available, higher resolutions (e.g. 300 dpi) greatly improve performance.

\begin{table}
\centering
\caption{Performance of WD classifiers on GPDS-300, using 12 reference signatures (Errors and standard deviations in \%)}
\label{tbl:wdperf}
\resizebox{\columnwidth}{!}{%
\ra{1.5}
\begin{tabular}{llll}
\toprule
Feature         & \specialcell{Training \\ Algorithm}  & EER\textsubscript{global threshold} & EER\textsubscript{user thresholds} \\
\midrule
SigNet-300dpi &  & 5.72 ($\pm 0.21$) &    3.5 ($\pm 0.22$) \\
SigNet-SPP-300dpi& Fixed &  5.63 ($\pm 0.22$) &   3.15 ($\pm 0.14$) $\bullet$ \\
SigNet-SPP-300dpi& Multi &  7.75 ($\pm 0.28$) &   4.86 ($\pm 0.24$)  \\ \midrule
SigNet-300dpi-F &&  1.78 ($\pm 0.12$) &    0.4 ($\pm 0.08$)  \\
SigNet-SPP-300dpi-F & Fixed &   1.69 ($\pm 0.1$) &   0.41 ($\pm 0.05$)  \\
SigNet-SPP-300dpi-F& Multi &  2.52 ($\pm 0.09$) &    0.8 ($\pm 0.07$) \\ \midrule
SigNet-600dpi &  &7.11 ($\pm 0.17$) &    4.2 ($\pm 0.27$) \\
SigNet-SPP-600dpi & Fixed&  7.06 ($\pm 0.13$) &   4.02 ($\pm 0.18$) \\
SigNet-SPP-600dpi & Multi &  6.36 ($\pm 0.16$) &   3.96 ($\pm 0.23$) \\ \midrule
SigNet-600dpi-F & & 2.46 ($\pm 0.09$) &    0.8 ($\pm 0.08$) \\
SigNet-SPP-600dpi-F & Fixed &  2.27 ($\pm 0.18$) &   0.65 ($\pm 0.11$) $\bullet$ \\
SigNet-SPP-600dpi-F & Multi&  2.85 ($\pm 0.16$) &    0.86 ($\pm 0.1$) \\
\bottomrule
\end{tabular}

}
\end{table}

We now consider the experiments using SPP for learning a fixed-sized representation for signatures of different sizes. Table \ref{tbl:wdperf} compares the performance of the WD classifiers on the GPDS dataset, as we change the training method. We consider both the baseline (network without SPP), and the two proposed training protocols for using SPP: with a single fixed canvas for training (denoted ``Fixed" in the table), and using the 5 different canvases, defined in the development set (denoted ``Multi" in the table). The results that are significantly better than the baseline (at $p < 0.01$) are denoted with a bullet point ($\bullet$). We notice that the performance between the baseline and SPP Fixed is very similar, while the method using multiple canvases during training performs a little worse. The proposed method using SPP Fixed keeps about the same level of performance as the baseline, while removing the constraint of having a maximum signature size (since both SPP methods accept larger signatures for processing).

\begin{figure}
\centering
\includegraphics[width=\columnwidth]{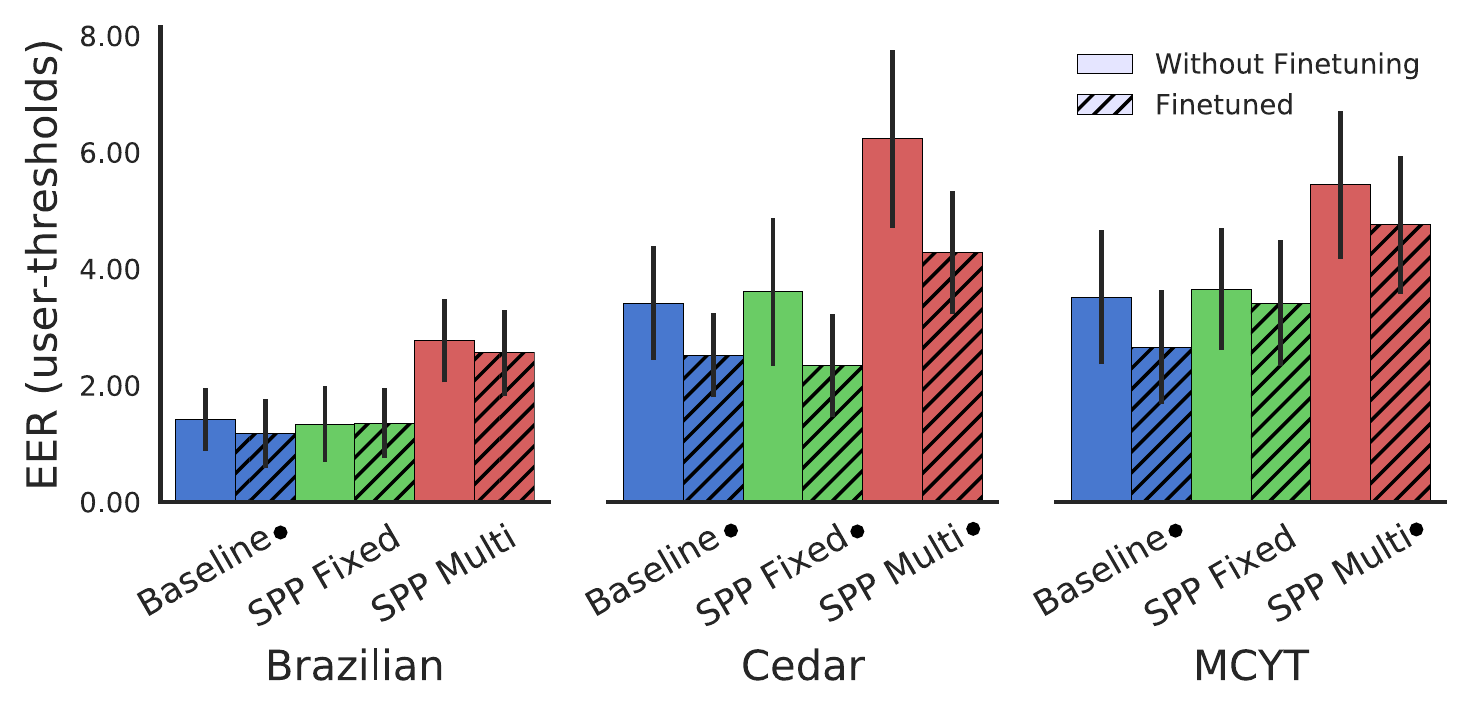}
\caption{Classification performance of Writer-Dependent classifiers trained with representations learned in the GPDS dataset. The hatched bars denote results with features fine-tuned in each particular dataset. Error bars denote the standard deviation across 100 replications.}
\label{fig:transfer_noforg}
\end{figure}

The results on transferring representations to different operating conditions are summarized in figure \ref{fig:transfer_noforg}. We considered models trained on the GPDS dataset, and used these models to extract features and train WD classifiers on other operating conditions, that is, three other datasets: Brazilian PUC-PR, Cedar and MCYT.  In all cases, we verify the impact of fine-tuning the representations for the new operating conditions, following the procedure detailed in section \ref{sec:finetuning}. For the first two datasets, that were scanned in 300 dpi, we used the representations learned in GPDS at 300 dpi, while for MCYT we used the representations learned at 600 dpi. In this experiment, we did not use any forgeries for training (neither from GPDS nor the target dataset). We performed t-tests to verify if fine-tuning the representation significantly improved the classification performance (marked with a bullet point next to the dataset name). We can see that the baseline (without SPP) and the SPP model trained with fixed image sizes performed similarly, while SPP trained on multiple canvas sizes performed worse for transfer. We also consistently see that fine-tuning representations on the target datasets helps the domain adaptation, reducing the errors on average.

\begin{table}
	\caption{Generalization performance on other datasets, with and without fine-tuning (for \emph{random forgeries}) }
	\label{tbl:results_otherscripts}
	\resizebox{\columnwidth}{!}{%
	\begin{tabular}{llll}
		\toprule
		Dataset & Finetuned &             EER\textsubscript{global threshold} &            EER\textsubscript{user thresholds}  \\
		\midrule
		Bengali &     &   5.07 ($\pm 0.8$) &  3.41 ($\pm 0.81$)  \\
		Bengali &     Yes &  0.77 ($\pm 0.27$)  &  0.16 ($\pm 0.14$) $\bullet$  \\ \midrule
		Devanagari &       &  4.65 ($\pm 0.92$) &   2.93 ($\pm 0.8$)  \\
		Devanagari &      Yes &   0.33 ($\pm 0.2$)  &  0.06 ($\pm 0.09$) $\bullet$ \\ \midrule
		MCYT &       &  0.19 ($\pm 0.39$) &  0.03 ($\pm 0.13$)  \\
		MCYT &       Yes &  0.04 ($\pm 0.12$)  &    0.0 ($\pm 0.0$)  \\ \midrule
		CEDAR &       &  1.14 ($\pm 0.75$) &  0.37 ($\pm 0.42$) \\
		CEDAR &       Yes &  0.23 ($\pm 0.26$)  &  0.08 ($\pm 0.19$) $\bullet$ \\ \midrule
		Brazilian &      &   0.47 ($\pm 0.3$) &   0.2 ($\pm 0.25$)\\
		Brazilian &       Yes &   0.5 ($\pm 0.38$) &  0.16 ($\pm 0.24$)  \\ \midrule \midrule
		\specialcell{Bengali \\ (Results from  \cite{ferrer_static_2017})}
		&      - &   0.67 &  - \\
		\specialcell{Devanagari \\ (Results from  \cite{ferrer_static_2017})} &       - &  0.47 &  - \\ 
		
		\bottomrule
		
	\end{tabular}
}
	
\end{table}

Table \ref{tbl:results_otherscripts} shows the results of the experiments on transferring the representation to other types of scripts. The objective of this experiment was to verify if the features learned on the GPDS dataset generalizes to other types of script (in particular, we tested for Bengali and Devanagari). Differently from the previous analysis, for these datasets we consider the performance on discriminating genuine signatures and random forgeries (signatures from other users), since skilled forgeries are not available in these synthetic datasets. We consider experiments using the network trained on GPDS with no changes, and experiments where we finetune the representation to the particular dataset, following the protocol from section \ref{sec:finetuning}. Results that are significantly better (at $p < 0.01$) are shown with a bullet. We noticed an interesting trend in these results, where without finetuning, the performance on other datasets that contain western-style signatures is already good (around or less than 1\% for MCYT, CEDAR and Brazilian PUC-PR), but for the indian scripts the performance was much worse (3-5\% EER). By finetuning the representation for the scripts, we obtain good performance (comparable with the previously reported in \cite{ferrer_static_2017}). This suggests that the learned representation generalize better to users with western scripts than to other scripts. Multi-script learning approaches could be considered to improve performance on all scripts.

\begin{table}
\centering
\caption{Comparison with state-of-the art on the GPDS dataset (errors in \%)}
\label{tbl:soa_gpds}

\resizebox{\columnwidth}{!}{%
\begin{threeparttable}
\ra{1.5}

\begin{tabular}{llllr}
\toprule
 Reference & Dataset& \begin{tabular}[x]{@{}c@{}}\#samples\\per user\end{tabular}  &Features &  EER\\
\midrule
%Vargas et al \cite{vargas_off-line_2010}  &GPDS-100 & 5 &Wavelets & 14.22 \\
%Vargas et al \cite{vargas_off-line_2011} & GPDS-100 &10 &LBP, GLCM&  9.02 \\

Hu and Chen \cite{hu_offline_2013}& GPDS-150 &10 &LBP, GLCM, HOG & 7.66\\
Guerbai et al \cite{guerbai_effective_2015} & GPDS-160&12&Curvelet transform&15.07\\
Serdouk et al \cite{serdouk_new_2015} & GPDS-100 & 16 & GLBP, LRF& 12.52 \\
Yilmaz \cite{yilmaz_score_2016}  &GPDS-160 &5&LBP, HOG, SIFT& 7.98\\
Yilmaz \cite{yilmaz_score_2016}  &GPDS-160 &12&LBP, HOG, SIFT& 6.97\\
Soleimani et al \cite{soleimani_deep_2016}  &GPDS-300 &10&LBP& 20.94\\
Hafemann et al \cite{hafemann_pr_2017}  & GPDS-300 & 12  & SigNet-F& 1.69 ($\pm 0.18$)\\
\midrule

Present Work  & GPDS-300 & 12  & SigNet-SPP-300dpi& 3.15 ($\pm0.14$)\\

Present Work  & GPDS-300 & 12  & SigNet-SPP-300dpi-F& 0.41 ($\pm0.05$)\\

\bottomrule
\end{tabular}

\end{threeparttable}
}
\end{table}

\begin{table}
\centering
\caption{Comparison with the state-of-the-art in MCYT (errors in \%)}
\label{tbl:mcyt_soa}
\resizebox{\columnwidth}{!}{%
\begin{tabular}{lrlc}
\toprule
Reference & \begin{tabular}[x]{@{}c@{}}\#samples\\per user\end{tabular} & Features & EER\\
\midrule
Gilperez et al.\cite{gilperez_off-line_2008}&5&Contours (chi squared distance)&10.18\\
Gilperez et al.\cite{gilperez_off-line_2008}&10&Contours (chi squared distance)&6.44\\
Wen et al.\cite{wen_model-based_2009}&5&RPF (HMM)&15.02\\
Vargas et al.\cite{vargas_off-line_2011}&5&LBP (SVM)&11.9\\
Vargas et al.\cite{vargas_off-line_2011}&10&LBP (SVM)&7.08\\
Ooi et al\cite{ooi_image-based_2016}&5&DRT + PCA (PNN)&13.86\\
Ooi et al\cite{ooi_image-based_2016}&10&DRT + PCA (PNN)&9.87\\
Soleimani et al.\cite{soleimani_deep_2016}&5&HOG (DMML)&13.44\\
Soleimani et al.\cite{soleimani_deep_2016}&10&HOG (DMML)&9.86\\
Hafemann et al \cite{hafemann_pr_2017} & 10 & SigNet (SVM) & 2.87 ($\pm$ 0.42) \\

\midrule

Present Work&10&SigNet-SPP-600dpi&3.64 ($\pm$ 1.04)\\
Present Work&10&SigNet-SPP-600dpi (finetuned)&3.40 ($\pm$ 1.08)\\

\bottomrule
\end{tabular}
}
\end{table}

\begin{table}
\centering
\caption{Comparison with the state-of-the-art in CEDAR (errors in \%)}
\label{tbl:cedar_soa}
\resizebox{\columnwidth}{!}{%

\begin{tabular}{lrlc}
\toprule
Reference & \begin{tabular}[x]{@{}c@{}}\#samples\\per user\end{tabular} & Features & AER/EER\\
\midrule
Chen and Srihari\cite{chen_new_2006}&16&Graph Matching&7.9\\
Kumar et al.\cite{kumar_writer-independent_2010}&1&morphology (SVM)&11.81\\
Kumar et al.\cite{kumar_writer-independent_2012}&1&Surroundness (NN)&8.33\\
Bharathi and Shekar\cite{bharathi_off-line_2013}&12&Chain code (SVM)&7.84\\
Guerbai et al.\cite{guerbai_effective_2015}&4&Curvelet transform (OC-SVM)&8.7\\
Guerbai et al.\cite{guerbai_effective_2015}&8&Curvelet transform (OC-SVM)&7.83\\
Guerbai et al.\cite{guerbai_effective_2015}&12&Curvelet transform (OC-SVM)&5.6\\
Hafemann et al. \cite{hafemann_pr_2017} & 12 & SigNet-F (SVM) & 4.63 ($\pm$ 0.42) \\
\midrule

Present Work&10&SigNet-SPP-300dpi&3.60 ($\pm$ 1.26)\\
Present Work&10&SigNet-SPP-300dpi (finetuned)&2.33 ($\pm$ 0.88)\\

\bottomrule
\end{tabular}
}
\end{table}

\begin{table*}
	\centering
	\caption{Comparison with the state-of-the-art on the Brazilian PUC-PR dataset (errors in \%)}
	\label{tbl:soa_brazilian}
	\ra{1.5}
	\resizebox{\textwidth}{!}{%
		
		\begin{tabular}{lllrrrrrrr}
			\toprule
			Reference & \begin{tabular}[x]{@{}c@{}}\#samples\\per user\end{tabular}  & Features &  FRR &  FAR\textsubscript{random} &  FAR\textsubscript{simple} &  FAR\textsubscript{skilled} &  AER & AER\textsubscript{genuine + skilled} &  EER\textsubscript{genuine + skilled}  \\
			\midrule
			Bertolini et al. \cite{bertolini_reducing_2010} & 15 &Graphometric & 10.16&3.16&2.8&6.48&5.65&8.32 & - \\
			Batista et al. \cite{batista_dynamic_2012} & 30 & Pixel density & 7.5&0.33&0.5&13.5&5.46&10.5 & -\\
			Rivard et al. \cite{rivard_multi-feature_2013} & 15 &ESC + DPDF &11&0&0.19&11.15&5.59&11.08& -\\
			Eskander et al. \cite{eskander_hybrid_2013}& 30 &ESC + DPDF &7.83&0.02&0.17&13.5&5.38&10.67& -\\
			Present Work &                15 &               SigNet &   1.22 ($\pm$ 0.63) &  0.02 ($\pm$ 0.05) &  0.43 ($\pm$ 0.09) &  10.70 ($\pm$ 0.39) &  3.09 ($\pm$ 0.20) &  5.96 ($\pm$ 0.40) &     2.07 ($\pm$ 0.63) \\
			\midrule
			
			Present Work & 15 & SigNet-SPP-300dpi    &  0.69 ($\pm 0.51$) &  0.04 ($\pm 0.07$) &  0.14 ($\pm 0.2$) &  9.51 ($\pm 1.27$) &  2.59 ($\pm 0.35$) &  5.1 ($\pm 0.69$) &   1.33 ($\pm 0.65$) \\
			Present Work & 15 & SigNet-SPP-300dpi (finetuned)   &   0.63 ($\pm 0.57$) &  0.03 ($\pm 0.07$) &  0.14 ($\pm 0.2$) &  8.78 ($\pm 1.55$) &  2.39 ($\pm 0.39$) &  4.7 ($\pm 0.77$) &    1.35 ($\pm 0.6$) \\

			\bottomrule
		\end{tabular}
	}
\end{table*}

Lastly, tables \ref{tbl:soa_gpds}, \ref{tbl:mcyt_soa}, \ref{tbl:cedar_soa} and \ref{tbl:soa_brazilian} compare the results we obtained with SPP-Fixed (considering EER\textsubscript{user thresholds} using genuine signatures and skilled forgeries) with the state-of-the-art in GPDS, MCYT, Cedar and Brazilian PUC-PR, respectively. We observe results competitive to the state of the art in all datasets. In particular, in the GPDS dataset we notice big gains in performance (0.41\% EER compared to 1.69\% EER).

It is also worth noting that the MCYT dataset contains both Offline and Online signature data (for the same users). Historically, performance on online systems was greatly superior, but recent work on offline signature verification is closing the gap between the two strategies. The best results on the literature achieve 2.85\% EER \cite{rua_online_2012} and 3.36\% EER \cite{fierrez_hmm-based_2007} on the Online MCYT dataset, while for offline signature verification, performance is achieving around 3\% EER\textsubscript{user thresholds}. Although these results are not directly comparable (both \cite{rua_online_2012} and \cite{fierrez_hmm-based_2007} implement per-user score normalization with a single global threshold), it shows that the gap between the two approaches is being reduced.

\section{Conclusion}

In this work we proposed and evaluated two methods for adapting the CNN architectures to learn a fixed-size representation for signatures of different sizes. A simple method, of training a network with SPP in images of a fixed sized (and generalizing to signatures of any size) showed similar performance to previous methods, while removing the constraint of having a maximum signature size that could be processed.

Our experiments with different resolutions showed that using larger image resolutions do not always lead to improved performance. In particular, we empirically showed that using resolution higher than 100 dpi greatly improves performance if skilled forgeries (from a subset of users) is used for feature learning, but does not improve performance if only genuine signatures are used. This suggests that when learning features from skilled forgeries, the network can use detailed information about the signature strokes (e.g. if the writing is shaky, with limp strokes), while this information is ignored when only genuine signatures are used for training the CNN (when the network is only attempting to distinguish between different users). 

Lastly, our experiments with transfer learning confirm previous results that features learned in one signature dataset generalize to other operating conditions. Our results also suggest that fine-tuning the representations (on a subset of the users in the new dataset) is useful to adapt the representations to the new conditions, improving performance. Especially for signatures from different styles than used for training (e.g. CNN trained on western signatures and generalizing to other types of script), finetuning showed to be particularly important. Other techniques, such as multi-script learning are also be promising for this scenario.

\begin{acknowledgements}

This work was supported by the Fonds de recherche du Qu\'ebec - Nature et technologies (FRQNT), the CNPq grant \#206318/2014-6 and by the grant RGPIN-2015-04490 to Robert Sabourin from the NSERC of Canada.  In addition, we gratefully acknowledge the support of NVIDIA with the donation of the Titan Xp GPU used in this research.

\end{acknowledgements}

% BibTeX users please use one of
\bibliographystyle{spbasic}      % basic style, author-year citations
%\bibliographystyle{spmpsci}      % mathematics and physical sciences
%\bibliographystyle{spphys}       % APS-like style for physics
%\bibliography{}   % name your BibTeX data base

\bibliography{biblio}

\end{document}